\documentclass[conference]{IEEEtran}
\IEEEoverridecommandlockouts
\usepackage{cite}
\usepackage{amsmath,amssymb,amsfonts}
\usepackage{algorithmic}
\usepackage[ruled]{algorithm2e}
\usepackage{graphicx}
\usepackage{float}
\usepackage{subfigure}
\usepackage{array}
\usepackage{textcomp}
\usepackage{booktabs}
\usepackage[colorlinks=false]{hyperref}
\usepackage{xcolor}
\usepackage{makecell}
\usepackage{arydshln}
\usepackage{cite}
\usepackage{dsfont}
\usepackage{bbm}

\usepackage{balance}
\usepackage{caption}
\makeatletter
\newcommand*{\rom}[1]{\expandafter\@slowromancap\romannumeral #1@}
\makeatother
\def\BibTeX{{\rm B\kern-.05em{\sc i\kern-.025em b}\kern-.08em
    T\kern-.1667em\lower.7ex\hbox{E}\kern-.125emX}}
\PassOptionsToPackage{bookmarks=false}{hyperref}
\pdfoutput=1

\begin{document}

\title{Skeleton-Based Action Recognition with Spatial-Structural Graph Convolution}
% \author{\IEEEauthorblockN{Anonymous Authors}}
\author{\IEEEauthorblockN{Jingyao WANG}
\IEEEauthorblockA{\textit{LIMOS, CNRS UMR 6158} \\ \textit{LPC, CNRS UMR 6533} \\
Université Clermont Auvergne \\jingyao.wang@uca.fr}\and\IEEEauthorblockN{Issam FALIH}\IEEEauthorblockA{\textit{LIMOS, CNRS UMR 6158} \\Université Clermont Auvergne
\\issam.falih@uca.fr}\and\IEEEauthorblockN{Emmanuel BERGERET }\IEEEauthorblockA{\textit{LPC, CNRS UMR 6533} \\Université Clermont Auvergne \\emmanuel.bergeret@uca.fr}
}
\maketitle

\begin{abstract}
\footnote{Code is available at: \url{https://github.com/jingyaojade/SpSt-GCN}} 
Human Activity Recognition (HAR) is a field of study that focuses on identifying and classifying human activities. Skeleton-based Human Activity Recognition has received much attention in recent years, where  Graph Convolutional Network (GCN) based method is widely used and has achieved remarkable results. However, the representation of skeleton data and the issue of over-smoothing in GCN still need to be studied. 1). Compared to central nodes, edge nodes can only aggregate limited neighbor information, and different edge nodes of the human body are always structurally related. However, the information from edge nodes is crucial for fine-grained activity recognition. 2). The Graph Convolutional Network suffers from a significant over-smoothing issue, causing nodes to become increasingly similar as the number of network layers increases.
Based on these two ideas, we propose a two-stream graph convolution method called Spatial-Structural GCN (SpSt-GCN). Spatial GCN performs information aggregation based on the topological structure of the human body, and structural GCN performs differentiation based on the similarity of edge node sequences. 
The spatial connection is fixed, and the human skeleton naturally maintains this topology regardless of the actions performed by humans. However, the structural connection is dynamic and depends on the type of movement the human body is performing. Based on this idea, we also propose an entirely data-driven structural connection, which greatly increases flexibility.  
We evaluate our method on two large-scale datasets, i.e., NTU RGB+D and NTU RGB+D 120. The proposed method achieves good results while being efficient. 
\end{abstract}


\section{Introduction}
\setlength{\parindent}{0em}

Human activity recognition (HAR) is a research field that spans computer science and electronic engineering. Its goal is to identify and classify human activities using algorithms. Human Activity Recognition (HAR) has received significant attention in recent years and has been applied in various fields such as healthcare, sports, security, smart homes, and wearable devices. The aim is to enhance human well-being, safety, and productivity by offering insights and information for decision-making, monitoring, or feedback purposes\cite{sun2022human}. The HAR process involves several steps: data collection, data preprocessing, feature extraction, and activity recognition\cite{sun2022human}. 

Various data modalities are used for different scenarios. They could be divided into two groups: visual modality and non-visual modality. RGB and skeleton data are two common types of visual modalities. RGB data is typically captured with a camera and contains a wealth of color and contour information. This data is widely used in security cameras, intelligent robots, and other applications. Skeleton data is usually extracted from RGB data or collected by depth cameras. Compared to RGB data, 3D skeleton data effectively captures spatial-temporal properties. It contains less data but has higher computational efficiency\cite{peng2021rethinking}. At the same time, it protects privacy. Skeleton data is a good choice when computing resources are limited or privacy protection is required. Acceleration data is typically non-visual data. The acceleration sensor is small, lightweight, and designed to protect privacy. It can accurately and comprehensively measure the spatial acceleration of an object without prior knowledge of its direction of motion. This sensor is commonly used in low-cost and low-power-consumption portable devices, such as smartwatches. Each of the various data modalities has its suitable application scenario. In this article, we focused on visual data, specifically on skeleton data. 

Currently, the predominant approach involves using spatial-temporal convolutional networks to extract features from skeleton data. Researchers utilize GCN to extract spatial features within frames, and then utilize TCN to extract temporal features between frames. The researcher designs the skeletal connection diagram based on the human skeleton's topology. When performing graph convolution, the adjacency matrix is utilized to depict the connections between nodes and perform message passing between neighboring joints\cite{yan2018spatial, song2022constructing}. The issue is that human beings have symmetrical structures, and there should be not only spatial connections but also structural connections between joints. 

The action of "clapping hands" requires both hands to come together. There is a strong connection between the two hands. However, a mere spatial connection would not adequately represent the connection between the hands, and we need to depict this connection with a structural link. We believe that distinguishing fine-grained activities requires a focus on the activities of edge nodes. Taking typing on the keyboard and writing as an example\cite{shahroudy2016ntu}, the most critical metric to distinguish between these two actions is the fine-grained movement of the hand. In terms of spatial structure, edge nodes are the least connected to other nodes and farthest away from the center node. This hinders their ability to effectively aggregate information from other nodes. To address this issue, we propose adding extra structural connections to the edge nodes.

The spatial connection is fixed, and the human skeleton naturally maintains this topology regardless of the actions performed by humans. However, the structural connection should be dynamic and depends on the type of movement the human body is performing. There is a strong connection between the hands when clapping, but not when checking the time (from watch) [19]. Based on this idea, we propose a data-driven method to initialize the structural connections based on the similarity of edge node sequences for various samples, so that each sample has its unique structural connection. 
GCN aggregates node information through edge connections to generate new node representations, which is effective but can lead to the problem of over-smoothing. Different from the normal GCN way of aggregating information, we propose performing node information differentiation through the structural branch. Our differentiation method can enhance the contrast between the edge nodes, and also reduce the issue of over-smoothing to some extent.

The main contributions of this paper are summarized as follows:
\begin{itemize}
    \item We propose a spatial-structural graph convolution method which can better represent symmetrical human structures and edge nodes.
    \item  Instead of initializing the adjacency matrix solely based on the count of adjacent nodes, we adopt a differential approach to initialize the structural connections of various samples. This differential initialization is contingent on the similarity of edge node sequences, enhancing the flexibility of graph convolution. %Instead of initializing the adjacency matrix based on the number of adjacent nodes, we differentially initialize the structural connections of different samples based on the similarity of edge node sequences, which makes graph convolution more flexible.
    \item To address the over-smoothing problem inherent in GCN, we utilize the inverse adjacency matrix to discern information pertaining to edge nodes.% We use the inverse adjacency matrix to differentiate edge node information instead of aggregating information, which helps alleviate the over-smoothing issue of GCN. 
\end{itemize}

The remainder of this paper is organized as follows: Section 2 presents the state of the art papers related to our work. Our proposed method is highlighted in section 3. Section 4 presents the experimental part including data set description, experimental settings, evaluation metrics, experimental results and their analysis and the conclusion is given in section 5.

\section{Related Works}

HAR is a classic temporal classification problem. In earlier years, traditional CNN and RNN models were dominant.
Zhang et al.\cite{zhang2019view} designed a two-stream CNN-RNN model to overcome view dependency. The main idea of this paper is to use view adaptation modeling to obtain a variable observation coordinate, allowing for the generation of skeleton data from different viewpoints. An RNN-based view adaptation subnetwork is integrated into the VA-RNN stream, while a CNN-based view adaptation subnetwork is integrated into the VA-CNN stream. Then, an LSTM classification network is applied in the VA-RNN stream, while a CNN classification network is applied in the VA-CNN stream. 
Caetano et al.\cite{caetano2019skelemotion} encoded the magnitude and orientation values of the skeleton joints into the image, and then utilized a CNN network for classification.

In recent years, several Transformer-based and GCN-based models have appeared.
Plizzari et al. \cite{plizzari2021spatial} developed a two-stream spatial-temporal transformer network. On the S-TR stream, a Spatial Self-Attention module (SSA) is used to analyze intra-frame interactions between different body parts, followed by a 2D convolution on the time dimension (TCN). On the T-TR stream, a Temporal Self-Attention module (TSA) is used to model inter-frame correlations, while spatial features are extracted by a standard graph convolution (GCN). 

Skeleton data used for human activity recognition (HAR) tasks are sequential data that contain consecutive frames. Skeleton data represents natural topography. In addition to being used as a matrix, skeleton data could also be utilized as a graph by a graph neural network \cite{scarselli2008graph}. In addition, the skeleton data used for action recognition is sequential data. GCN and TCN blocks are often used to capture spatial-temporal features, where Graph Convolutional Network (GCN) is used to capture intraframe spatial information by aggregating information from neighboring nodes, and Temporal Convolutional Network (TCN) is used to capture interframe temporal information.
Yan et al.\cite{yan2018spatial} propose the Spatial-Temporal Graph Convolutional Networks (ST-GCN) model, which applies alternating convolution of GCN and TCN, and has become the baseline in the field of skeleton-based action recognition. 
Song et al. \cite{song2022constructing} developed an efficient graph convolutional network. They extended the convolutional layer in CNN to the GCN network to extract temporal dynamics and compress the model size. They then employed a compound scaling method that adjusts the width (number of channels) and depth (number of layers) factors to make the model flexible and effective. In addition, Spatial Temporal Joint Attention (ST-JointAtt) is incorporated into each block, and three input branches are fused early in the network. The above strategies enable the network to achieve good results at a low computational cost.

\subsection{Skeleton Data Representation}

Skeleton data consists of a set of joint coordinates, usually represented as 3D points or 2D projections, along with additional information like joint velocities or orientations. These joint coordinates are often normalized to a consistent scale or relative to a specific reference point to enhance generalizability across individuals and improve recognition accuracy. Skeleton data removes video backgrounds, people textures and outlines, which reduces lots of calculations. 3D skeleton data has a good view-invariance. It is possible to change the data view by rotating the skeleton in 3D space. 

Some researchers use matrices or images to represent skeleton data. Yang et al. \cite{yang2018action} proposed a Tree Structure Skeleton Image (TSSI) which encodes skeleton data by depth-first tree traversal order to image. Banerjee et al. \cite{banerjee2020fuzzy} extracted four features from skeleton data (Distance feature, Distance velocity feature, Angle feature, Angle velocity feature). These features are then encoded into single-channel grayscale images and are fed to the CNN.

Skeleton data is a natural topography. In addition to being used as a matrix, skeleton data could also be represented as a graph. Some researchers represent skeleton data as an undirected graph, which effectively illustrates the connections between joints. The joint information will be aggregated bi-directionally. Yan et al.\cite{yan2018spatial} represented the skeleton data using natural bone topology based on prior experience. Instead of using a fixed graph, Shi et al. \cite{shi2019two} propose a data-driven method to enable the model to adapt to various data samples. According to the pattern of human movement, the central joints always drive the peripheral joints. Shi et al. \cite{shi2019skeleton} represented the skeleton as a directed acyclic graph to better incorporate joint and bone data. 

In addition to representations based on the original topology, some researchers utilize the Node2vec method to obtain graph embeddings. Laines et al. \cite{laines2023isolated} conducted a Depth-First-Search (DFS) on the skeleton tree to obtain a new spatial embedding.  Sun et al. \cite{sun2019cooperative} utilized the breadth-first search (BFS) algorithm, which commences from the root joint (hip center joint) to rearrange the joints.

In addition to utilizing fixed topological structures, some researchers have employed dynamic structures to represent the skeleton data. Ye et al.\cite{ye2020dynamic} proposed a Context-encoding Network (CeN) to automatically learn the skeleton topology, incorporating contextual features from the remaining joints in a global manner when learning the dependency between two joints. Cheng et al.\cite{cheng2020skeleton} proposed a non-local shift graph operation, which enables each node to have a receptive field covering the entire skeleton graph, allowing for adaptive learning of the relationships between joints. Chen et al.\cite{chen2021channel} proposed Channel-wise Topology Refinement Graph Convolution (CTR-GC), which learns a shared topology across all channels and refines it with channel-specific correlations for each channel.

\section{{Proposed Approach}}
In this work, we utilize the GCN-TCN block to extract spatial-temporal features, where the GCN layer and TCN layer are alternately employed to capture spatial and temporal dependencies. Temporal convolutions are essentially 2D convolutions with a kernel size of 1, designed to capture the temporal dependence of each node. In this section, we introduce our proposed spatial-structural graph convolution and the generation of the structural adjacency matrix. 

\subsection{Spatial-Structural Graph Convolution (SpSt-GCN)}
For GCN-based methods,  the input skeleton data is represented as (V, E) with N joints and T frames. Here, V represents joint points, and E represents connections between nodes, often represented by an adjacency matrix. 

The conventional spatial graph convolution is shown below, where $f_{in}$ and $f_{outSpatial}$ represent the input and output features, respectively. $A_{j}$ represents the adjacency matrix at distances j, $\Lambda_{j}$ is a weight vector used to normalize $A_{j}$. The parameters $W_{j}$ can be optimized during the training process. 
$$f_{outSpatial} = \sum_{j}W_{j}f_{in}\Lambda_{j}^{-\frac{1}{2}}A_{j}\Lambda_{j}^{-\frac{1}{2}}$$
To increase the flexibility of graph convolution, we add a parameterized matrix B as described in \cite{shi2019two}. The matrix B is initialized to an all-zero matrix of the same size as adjacency matrix A and optimized together with the other parameters in the training process.
$$f_{outSpatial} = \sum_{j}W_{j}f_{in}(\Lambda_{j}^{-\frac{1}{2}}A_{j}\Lambda_{j}^{-\frac{1}{2}}+B_{j})$$

Many researchers commonly employ the skeletal topology structure for establishing connections among joints, a methodology known for its efficiency albeit lacking in precision. It is crucial to recognize that the human skeleton, characterized by symmetry, presents a nuanced structure wherein even widely spaced symmetrical joints exhibit robust structural correlations. While prevailing graph convolution techniques generally enable non-edge nodes to efficiently aggregate information from both nearby and distant nodes, limitations arise when considering edge nodes. These edge nodes possess only a singular connection and, as a result, can solely aggregate information from nodes in close proximity to the central point. This asymmetry in information flow becomes particularly evident when comparing edge nodes to central nodes, as the former receives comparatively less neighbor information, thereby impeding the formation of optimal node representations. To address this challenge and enhance the representation of both symmetric and edge nodes, we propose a novel approach named spatial-structural graph convolution. 

\begin{figure}[htb]        
 \center{\includegraphics[width=8cm] {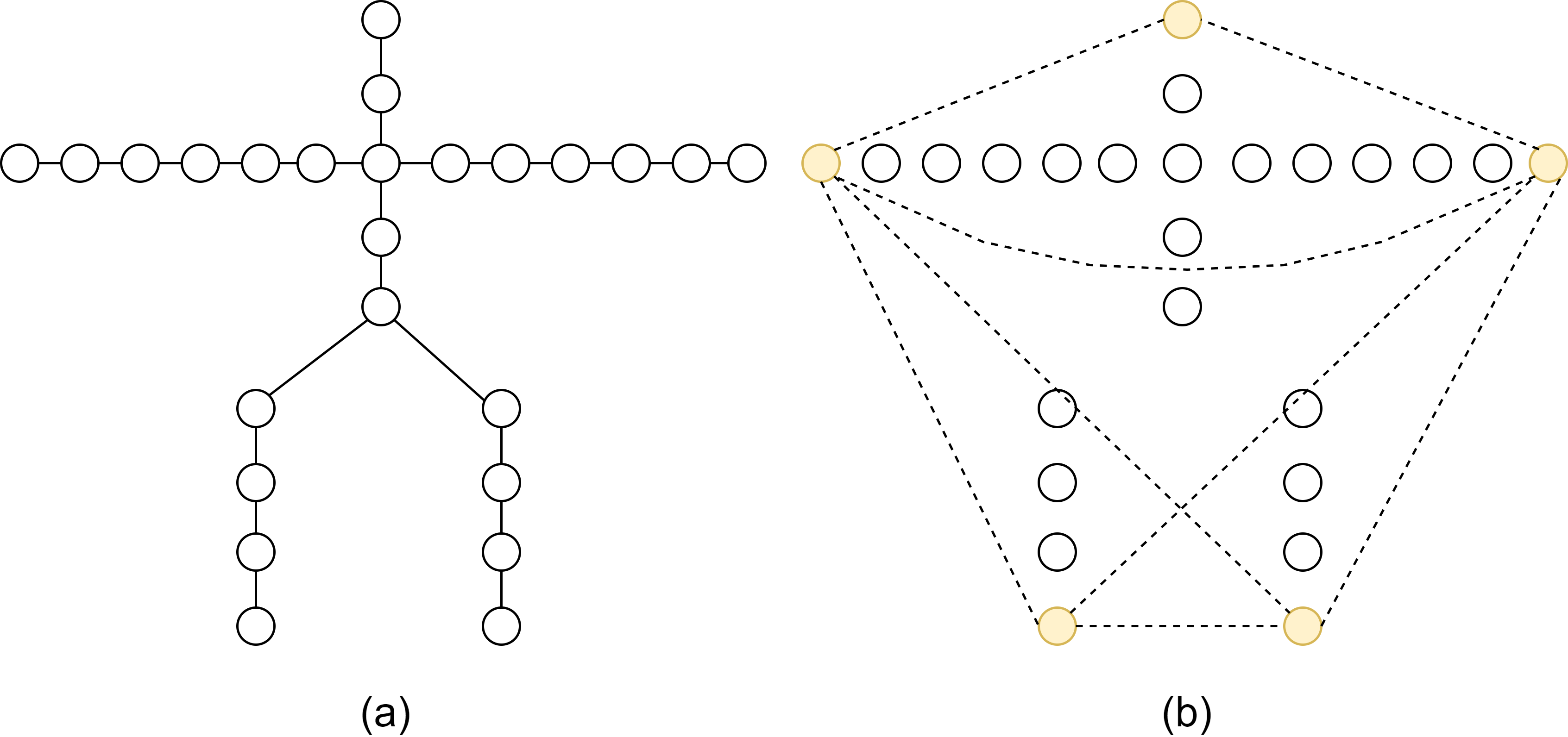}}        
 \caption{\label{graph representation} Graph representation of skeleton data. (a) is the graph representation for spatial graph convolution, (b) is the proposed graph representation for structural graph convolution.}     
\end{figure}

This innovative method involves the integration of two branches: a spatial branch responsible for aggregating information from spatially adjacent nodes and a structural branch focused on consolidating information between edge nodes. By combining these branches, the spatial-structural graph convolution aims to achieve a more comprehensive and accurate representation of the underlying skeletal structure, thereby advancing the state-of-the-art in joint connectivity modeling \ref{graph representation}.
The structural graph convolution is formulated as follows:
$$f_{outStructural} = \sum_{j}M_{j}f_{in}As_{j}$$
where $f_{in}$ and $f_{outStructural}$ represent the input and output features, respectively. $As_{j}$ represents the structural adjacency matrix for distance $j$. The parameters $M_{j}$ can be optimized during the training process. In this work, we set the distance as 1.  Then, we use element-wise addition to combine the outputs of these two branches and obtain the final output. 
$$f_{out} = f_{outSpatial}+f_{outStructural}$$

\subsection{Structural Adjacency Matrix}

In contrast to the conventional approach of spatial graph convolution, where all samples share a uniform adjacency matrix, we recognize that the structural connection between edge nodes is intricately linked with the specific actions performed. To capture these nuanced relationships, we propose a data-driven structural adjacency matrix, wherein the strength of connections between edge nodes is determined solely by the characteristics of each sample.

In the context of the NTU60 dataset's skeleton data, we observe five edge nodes, each corresponding to a distinct time series. In standard graph convolutional networks (GCN), the adjacency matrix is normalized based on the number of adjacent nodes. However, for structural connections, creating a conventional adjacency matrix between edge nodes is impractical as edge joints lack tangible spatial connections. Aggregating them without differentiation would compromise the unique characteristics of each edge node. Based on this idea, we use the similarity of edge node sequences instead of the number of adjacent nodes to represent the strength of the connection.

Euclidean distance and cosine similarity are widely employed distance calculation methods in deep learning. Euclidean distance quantifies the straight-line separation between two points in Euclidean space, making it well-suited for sequences with consistent lengths and where element order is crucial. Cosine similarity gauges the cosine of the angle between two vectors, treating sequences as vectors in a high-dimensional space. Cosine similarity is suitable for situations involving text data, document comparison, and situations where vector magnitudes are not critical.

In the context of skeleton data, the interplay of human muscles and bones results in a strong correlation among joint movements. However, the execution of force progresses incrementally. Consider playing badminton as an example: the upper arm propels the forearm, the forearm propels the wrist, leading to the final stroke of the shuttlecock. The transmission of the movement unfolds from the core to the periphery, with movement speed and amplitude gradually accelerating. Despite the collective coordination of the entire arm for a given motion, the time and range of movement for different joint points vary. In such cases, neither Euclidean distance nor cosine similarity proves suitable.

Taking inspiration from similarity calculation methods in natural language processing, we turn to the utilization of Dynamic Time Warping (DTW) to compute distances between each pair of edge nodes. DTW is a technique adept at measuring the similarity between two sequences that may exhibit variations in time or speed. It excels in scenarios where sequences share similar patterns but may be temporally misaligned. The DTW algorithm constructs an optimal warping path between the sequences, assessing similarity between corresponding points and allowing for time axis stretching or compressing as needed. The primary objective is to minimize the overall cost of warping while aligning similar features. Notably, DTW accommodates discrepancies in movement time among different joint points and considers the magnitude of movement, making it highly suitable for assessing movement similarity between edge nodes.

We finally use FastDTW for efficiency. We calculate a distance matrix $D$ for each sample. To discern the correlation between different edge nodes, we leverage the inverse of the distance matrix $D^{-1}$ as a representation of the correlation matrix. Larger values in $D^{-1}$ signify greater similarity in actions between the two edge nodes, indicating a stronger correlation.

Addressing the over-smoothing issue inherent in GCN, where data from different joints tends to become overly similar with each aggregation step, structural GCN takes a different approach. When employing the D-1 approach directly, heightened similarity among adjacent nodes leads to an aggregation of information, thereby exacerbating the likeness of edge nodes, a condition contrary to the desired outcome. Rather than engaging in information aggregation from neighboring nodes, this method prioritizes the preservation of node-specific information by delineating each edge node from others through the utilization of -D-1. Consequently, the application of -D-1 fosters greater differentiation among edge nodes, thereby facilitating the capture of nuanced information.
To achieve this, an identity matrix (I) is introduced to retain the distinct characteristics of each node. The dynamic structural adjacency matrix, denoted as $I-D^{-1}$, is then calculated for each sample, providing a nuanced representation of the structural connections within the data.

\begin{figure}[t]
\centering  %图片全局居中
\subfigure[throw]{
\includegraphics[height = 4cm]{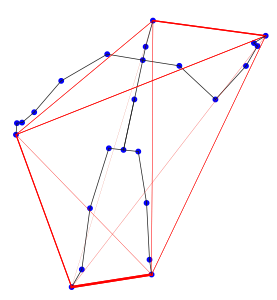}}\subfigure[writing]{
\includegraphics[height = 4cm]{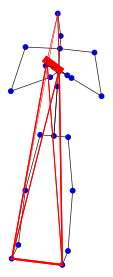}}\subfigure[stand up]{
\includegraphics[height = 4cm]{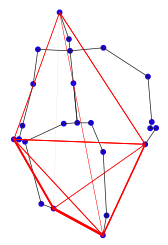}}
\caption{Spatial-Structural connection visualisation for joint input of NTU RGB+D dataset. The black line represents spatial connection and the red line represents structural connection. The thickness of the red line represents the strength of the structural connection.}
\label{visualization of skeleton}
\end{figure}

\begin{figure}[t]
\centering  %图片全局居中
\includegraphics[height = 3.2cm]{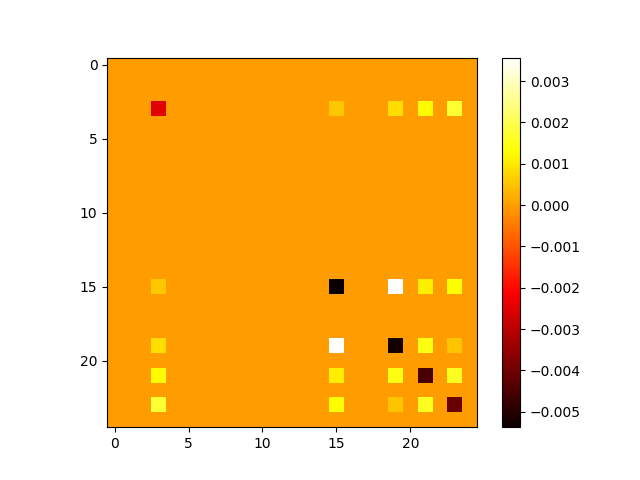}
\includegraphics[height = 3.2cm]{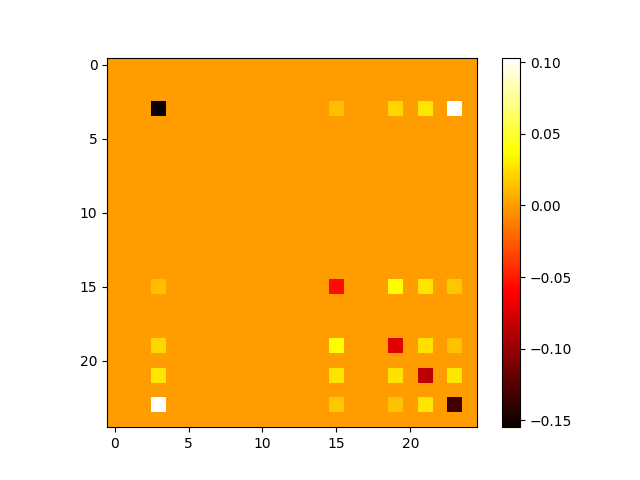}
\caption{Two examples of visualization of similarity matrix $D^{-1}$}
\label{visualization of structural adjacency matrix}
\end{figure}

\begin{figure}[t]
\centering  %图片全局居中
\includegraphics[height = 3.2cm]{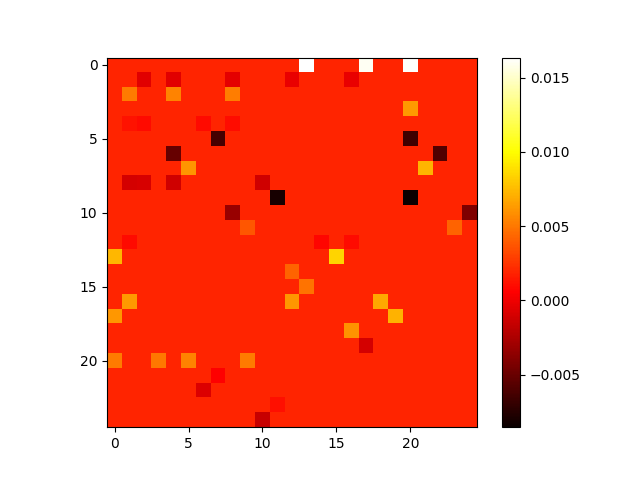}
\includegraphics[height = 3.2cm]{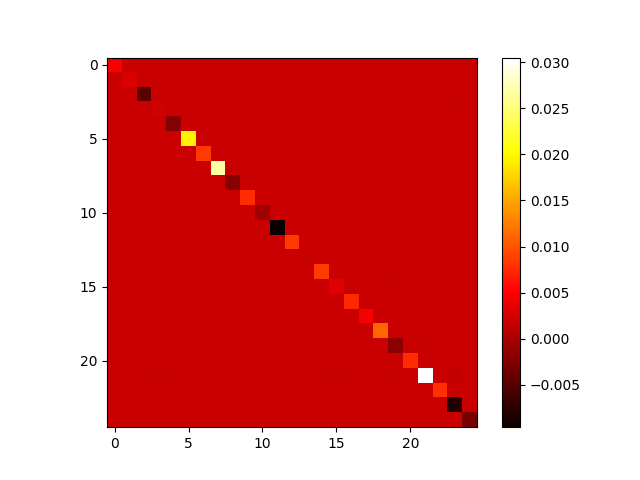}
\caption{Visualization of learned adjacency matrix. The left matrix is a part of the learned spatial matrix. The right matrix is an example of learned structural matrix.}
\label{visualization of learned adjacency matrix}
\end{figure}

For a visual representation, we refer to the algorithm in \ref{algo:event}, illustrating the steps involved in the dynamic structural adjacency matrix calculation. Additionally, Figure \ref{visualization of skeleton} provides visualizations of the spatial-structural connections for three distinct samples—throw, writing, and stand up—underscoring the effectiveness of our proposed approach in capturing the intricacies of fine-grained activity recognition. Figure \ref{visualization of structural adjacency matrix} provides visualizations of matrix $D^{-1}$. Figure \ref{visualization of learned adjacency matrix} provides visualizations of learned spatial and structural adjacency matrix.

\begin{algorithm}[t]
\caption{Structural Adjacency Matrix Calculation}
\label{algo:event}
\LinesNumbered
\KwIn{input sequence $i[C_{in},T_{in},V_{in}]$}
\KwOut{structural adjacency matrix $As$ with size $[V_{in}, V_{in}]$}
$D$: an all-zero array with size $[V_{in}, V_{in}]$;\\
$I$: an identity array with size $[V_{in}, V_{in}]$;\\
$edge$: edge nodes set;\\
\ForEach{i}{
    \For{a in edge}{
        \For{b in edge}{
            \uIf{$a \neq b \land D[a][b]==0$}{
                D[a][b] = \text{FastDTW}(i[:,:,a], i[:,:,b]);\\
                D[b][a] = D[a][b]\;
            }
        }
    }
$As = -D^{-1} + I$\;
}
\end{algorithm}

\subsection{Data Preprocessing}
In this work, we adopt the same data preprocessing as described in \cite{song2022constructing}. The shape of each action sequence $x$ is ($C_{in},T_{in},V_{in}$), where $C_{in}$ denotes coordinates , $T_{in}$ denotes frames, and $V_{in}$ denotes joints. We extract three input features from the input data: joint features, velocity features, and bone features. 

For joint features, we first calculate the relative joint positions $j_r$ of all joints $i$ ($i\in \left\{1,2,...,V_{in}\right\}$) to the center joint $c$, then concatenate the original joint positions with the relative joint positions. For the NTU RGB+D Dataset and NTU RGB+D 120 Dataset, we selected joint 2 (middle of the spine) as the central joint. Then we concatenate the original joint positions $x$  with the relative joint positions $j_r$ as joint input features.
$$j_r = x[:,:,i]-x[:,:,c]$$
$$j = x \oplus jr$$
For velocity features, we calculate slow motion $v_s$ for time $t$ ($t\in \left\{1,2,...,V_{in}\right\}$) using one frame as the time interval, and fast motion $v_f$ for time $t$ ($t\in \left\{1,2,...,V_{in}\right\}$)  using a two-frame time interval.  Then we concatenate the slow motion $v_s$  with the fast motion $v_f$ as velocity input features.
$$v_s = x[:,t+1,:]-x[:,t,:]$$
$$v_f = x[:,t+2,:]-x[:,t,:]$$
$$v = v_s \oplus v_f$$
For bone features, we first calculate the lengths of the bones $b_l$, which consist by joint $i$ ($i\in \left\{1,2,...,V_{in}\right\}$) and its adjacent joint $i_{adj}$ ($i_{adj}\in \left\{1,2,...,V_{in}\right\}$) . Then we calculate bones' angles $b_{a,w}$ for three coordinates $w$ ($w\in \left\{x,y,z\right\}$). We concatenate the lengths of the bones $b_l$ with the bone angles $b_{a,w}$ as bone input features.
$$b_l = x[:,:,i]-x[:,:,i_{adj}]$$
$$b_{a,w} = arccos(\frac{b_{l,w}}{\sqrt{b_{l,x}^{2}+b_{l,y}^{2}+b_{l,z}^{2}}})$$
$$b = b_l \oplus b_a$$

\subsection{Model}

\begin{figure*}[!htb]        
 \center        
 \includegraphics[width=0.9\textwidth]{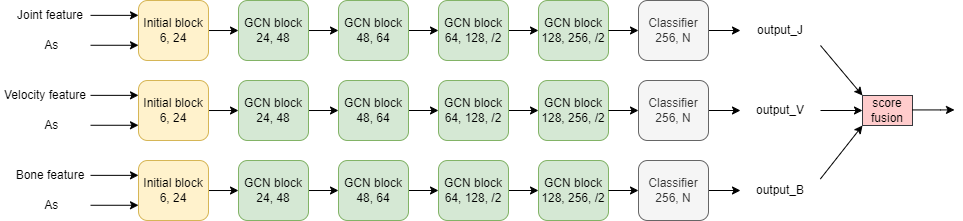}
 \caption{\label{model} Model structure, where N is the number of action classes, the numbers in the block represent the number of input channels and output channels, /2 represents a stride of 2}     
\end{figure*}

\begin{figure}[htb]
 \center{\includegraphics[width=8cm] {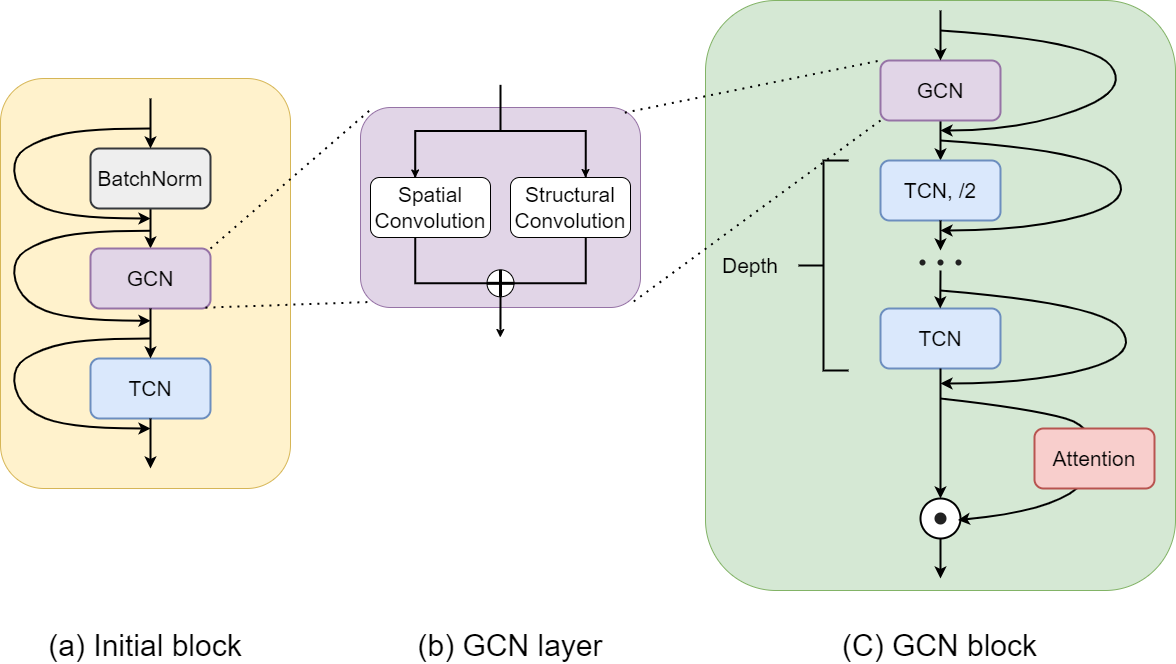}}        
 \caption{\label{block} Visualisation of Initial block, GCN layer and GCN block, where $\oplus$ and $\odot$ represent element-wise addition and element-wise product, respectively.}     
\end{figure}

For the overall efficiency of the network structure, we utilize the GCN block introduced by \cite{song2022constructing} as the fundamental unit \ref{block}. However, instead of the conventional GCN layer, we have integrated our novel spatial-structural graph convolution layer. The network architecture begins with an initial block designed to extract features at the outset, followed by the stacking of four GCN blocks to progressively extract higher-dimensional features. Finally, a Classifier block, which incorporates a Global Average Pooling (GAP) layer, a Dropout layer, and a Fully Connected (FC) layer, is employed to generate the output for each modality, as illustrated in \ref{model}. We then combine the results from three inputs (joint features, velocity features, bone features) to get the final output, where leveraging multiple modalities to enhance the model's capacity for nuanced feature extraction and robust predictive performance.

\section{Evaluation}

\subsection{Dataset}

\subsubsection{NTU RGB+D Dataset}
"NTU RGB+D" dataset\cite{shahroudy2016ntu} contains 60 action classes and 56,880 video samples captured from 40 different people using the Microsoft Kinect v2. The 60 action classes can be categorized into three main groups: 40 daily actions, 9 health-related actions, and 11 mutual actions. As mentioned, the NTU RGB+D dataset is a multi-modal dataset that contains RGB videos, depth map sequences, 3D skeletal data, and infrared (IR) videos for each sample. In this work, we only utilize skeleton data. Each frame contains two skeletons, and each skeleton contains 25 joints. 

We evaluated our method using two benchmarks proposed by the authors of this dataset: cross-subject validation (X-sub) and cross-view validation (X-view). In cross-subject validation, we divided the subjects into two groups: twenty subjects in the training data group and another twenty subjects in the validation group. In cross-view validation, the data collected by cameras 2 and 3 are recognized as training data, while the data collected by camera 1 is used for validation. 

\subsubsection{NTU RGB+D 120 Dataset}
The "NTU RGB+D 120" dataset\cite{liu2019ntu} is an extension of the "NTU RGB+D" dataset, captured from 106 different individuals, with an additional 60 classes and 57,600 video samples. The 120 action classes can be categorized into three main groups: 82 daily actions, 12 health-related actions, and 26 mutual actions. We evaluate our method using two benchmarks proposed by the authors of this dataset: cross-subject validation (X-sub) and cross-setup (X-set120). In cross-subject validation, 53 subjects are divided into a training data group and another 53 subjects into a validation group. In cross-validation, 16 setups are used for training, and the remaining 16 setups are used for testing.

\subsection{Experimental Setting}
In our experiments, we set the training epoch to 50. We apply the same settings as described in\cite{song2022constructing}. Our models are trained with gradient descent (SGD) with a Nesterov momentum of 0.9 and a weight decay of 0.0001. The learning rate is set to 0.1 and decays with a cosine schedule after the 10th epoch. The probability in the dropout layer is set to 0.25. We perform a data transformation as described in \cite{shi2019two} for the X-view benchmark. The batch size is set to 16 for both the NTU RGB+D dataset and the NTU RGB+D 120 dataset. All our experiments are performed on one NVIDIA GeForce RTX 3090 GPU.

\subsection{Comparisons}

\subsubsection{Effect of Structural Graph Convolution}
 In this part, we will discuss the impact of three input modalities and the impact of our structural branch on X-sub benchmark \ref{table:ss_ntu60} and on X-view benchmark \ref{table:ss_ntu120}. We can see that all three modalities are essential, and that fusing the scores of the three modalities outputs in the highest accuracy for both X-sub and X-view benchmarks. In addition, using the same network structure, there is a significant improvement in accuracy when employing the spatial-structural graph convolution layer instead of the spatial graph convolution layer.

\begin{table}[h]
  \centering
  \begin{tabular}{lcc}
      \toprule 
       Inputs & Spatial GCN & Spatial-Structural GCN \\
      \midrule
       Joint &  86.89 & \textbf{87.88} \\
       Velocity & 87.17 & \textbf{87.29} \\
       Bone & 87.82 & \textbf{88.20} \\
      \midrule
       Joint+Velocity & 90.19 & \textbf{90.69} \\
       Joint+Bone & 89.53 & \textbf{90.02} \\
       Velocity+Bone & 90.66 & \textbf{91.03} \\
      \midrule
       Joint+Velocity+Bone & 91.10 & \textbf{91.62} \\
      \bottomrule
  \end{tabular}
  \caption{Comparisons of different inputs on NTU60 X-sub benchmark in accuracy (\%)}
    \label{table:ss_ntu60}
\end{table}

\begin{table}[h]
  \centering
  \begin{tabular}{lcc}
      \toprule 
       Inputs & Spatial GCN & Spatial-Structural GCN \\
      \midrule
       Joint &  91.82 & \textbf{92.35} \\
       Velocity & 93.27 & \textbf{93.53} \\
       Bone & 92.27 & \textbf{92.94} \\
      \midrule
       Joint+Velocity & 95.08 & \textbf{95.23} \\
       Joint+Bone & 94.10 & \textbf{94.47} \\
       Velocity+Bone & 95.28 & \textbf{95.51} \\
      \midrule
       Joint+Velocity+Bone & 95.61 & \textbf{95.79} \\
      \bottomrule
  \end{tabular}
  \caption{Comparisons of different inputs on NTU60 X-view benchmark in accuracy (\%)}
    \label{table:ss_ntu120}
\end{table}

\subsubsection{NTU RGB+D Dataset}
We compare our models with the state-of-the-art methods on the NTU RGB+D dataset in \ref{ntu60}. We classify the commonly used methods for skeleton-based human activity recognition into two categories: CNN-RNN-LSTM-based methods and GCN-based methods. Notably, the GCN-based method exhibits superior performance compared to CNN-RNN-LSTM-based methods. In recent years, there has been a growing preference among researchers for GCN-based approaches in skeleton recognition tasks, showcasing better overall performance than their CNN-RNN-LSTM counterparts. 
ST-GCN is one of the most well-known GCN-based models for skeleton recognition. Our method greatly exceeds it by 10.1\% on the X-sub benchmark and 7.5\% on the X-view benchmark. EfficientGCN is another popular model for its high efficiency and high accuracy. In this work, we use the same GCN block and the same number of blocks as EfficientGCN-B0. The accuracy of our model exceeds it by 1.4\% on the X-sub bencnmark and 0.9\% on the X-view benchmark. CTR-GCN learns topologies dynamically, which demonstrates commendable outcomes. However, it overlooks critical considerations such as the symmetry inherent in human body structure and the constraints associated with edge nodes. We posit that integrating the adaptable topology of CTR-GCN with our refined graph representation will likely yield superior results. We consider that the results of StSp-GCN exceed those of most models due to our precise expression of the skeleton data and the flexibility of graph convolution by introducing unique structural connections for each sample.

\begin{table}[h]
   \centering
  \begin{tabular}{llcc}
      \toprule 
      Model & Conference &  X\_sub & X\_view \\
      \midrule
      ST-LSTM\cite{liu2016spatio} & ECCV16 & 69.2 & 77.7 \\
      STA-LSTM\cite{song2017end} & AAAI17 & 73.4 & 81.2 \\
      HCN\cite{li2018co} & IJCAI18 & 86.5 & 91.1 \\
      VA-fusion\cite{zhang2017view} & TPAMI19 & 89.4 & 95.0 \\
      \midrule
      ST-GCN\cite{yan2018spatial} & AAAI18 & 81.5 & 88.3 \\
      SR-TSL\cite{si2018skeleton} & ECCV18 & 84.8 & 92.4 \\
      AS-GCN\cite{li2019actional} & CVPR19 & 86.8 & 94.2 \\
      2s-AGCN\cite{shi2019two} & CVPR19 & 88.5 & 95.1 \\
      AGC-LSTM\cite{si2019attention} & CVPR19 & 89.2 & 95.0 \\
      DGNN\cite{shi2019skeleton} & CVPR19 & 89.9 & 96.1 \\
      SGN\cite{zhang2020semantics} & CVPR20 & 89.0 & 94.5 \\
      PL-GCN\cite{huang2020part} & AAAI20 & 89.2 & 95.0 \\
      NAS-GCN\cite{peng2020learning} & AAAI20 & 89.4 & 95.7 \\
      4s-Shift-GCN\cite{cheng2020skeleton} & CVPR20 & 90.7 & 96.5 \\
      DC-GCN+ADG\cite{cheng2020decoupling} & ECCV20 & 90.8 & 96.6 \\
      Dynamic-GCN\cite{ye2020dynamic} & ACMMM20 & 91.5 & 96.0 \\
      ST-TR\cite{plizzari2021spatial} & PR21 & 89.9 & 96.1 \\
      EfficientGCN-B0\cite{song2022constructing} & TPAMI21 & 90.2 & 94.9 \\
      CTR-GCN\cite{chen2021channel} & CVPR21  & 92.4  & 96.8 \\
      \midrule
      SpSt-GCN(ours)& & 91.6&  95.8\\
      \bottomrule
  \end{tabular}
  \caption{Comparisons with SOTA methods on NTU RGB+D dataset in accuracy (\%)}
  \label{ntu60}
\end{table}

\subsubsection{NTU RGB+D 120 Dataset}

Compared with the NTU RGB+D data set, the NTU RGB+D 120 dataset is larger and has more classes. This means that there are more similar actions that need to be distinguished, which poses a greater challenge to the model. The overall performance of the GCN-based model is still better than the CNN-RNN-LSTM-based method. Compared to ST-GCN, our method greatly exceeds it by 17.1\% on the X-sub benchmark and 15.6\% on the X-view benchmark. Our method outperforms EfficientGCN-B0 by 1.2\% on the X-sub benchmark and 3.8\% on the X-view benchmark. In comparison to CTR-GCN, our introduced structural connections augment the delineation of topology across diverse samples. It is our conjecture that amalgamating our structural links with the inherently flexible topology of CTR will yield more favorable outcomes. Similar to the results on the NTU RGB+D dataset, the performance of SpSt-GCN is mainly attributed to the precise expression of the skeleton data and the flexibility of graph convolution.

\begin{table}[h]
\centering
  \begin{tabular}{llcc}
      \toprule
      Model & Conference &  X\_sub & X\_view \\
      \midrule 
      PA-LSTM\cite{shahroudy2016ntu} & CVPR16 & 25.5 & 26.3 \\
      ST-LSTM\cite{liu2016spatio} & ECCV16 & 55.7 & 57.9 \\
      FSNet\cite{liu2019skeleton} & TPAMI19 & 59.9 & 62.4 \\
      \midrule
      ST-GCN*\cite{yan2018spatial} & AAAI18 & 70.7 & 73.2 \\
      SR-TSL*\cite{si2018skeleton} & ECCV18 & 74.1 & 79.9 \\
      AS-GCN*\cite{li2019actional} & CVPR19 & 77.9 & 78.5 \\
      2s-AGCN*\cite{shi2019two} & CVPR19 & 82.5 & 84.2 \\
      SGN\cite{zhang2020semantics} & CVPR20 & 79.2 & 81.5 \\
      4s-Shift-GCN\cite{cheng2020skeleton} & CVPR20 & 85.9 & 87.6 \\
      DC-GCN+ADG\cite{cheng2020decoupling} & ECCV20 & 86.5 & 88.1 \\
      MS-G3D\cite{liu2020disentangling} & CVPR20 & 86.9 & 88.4 \\
      Dynamic-GCN\cite{ye2020dynamic} & ACMMM20 & 87.3 & 88.6 \\
      ST-TR\cite{plizzari2021spatial} & PR21 & 81.9 & 84.1 \\
      EfficientGCN-B0\cite{song2022constructing} & TPAMI21 & 86.6 & 85.0 \\
      MST-GCN\cite{chen2021multi} & AAAI21 & 87.5 & 88.8 \\
      CTR-GCN\cite{chen2021channel} & CVPR21  & 88.9  & 90.6\\
      \midrule
      SpSt-GCN(ours)& & 87.8 & 88.8\\
      \bottomrule 
  \end{tabular} \\
  *: These results are implemented by \cite{song2022constructing}
  \caption{Comparisons with SOTA methods on NTU RGB+D 120 dataset in accuracy (\%)}
  
  \label{ntu120}
\end{table}

\subsubsection{Model Complexity}

Considering the importance of efficient resource utilization, we not only assess model accuracy but also consider computational complexity. We compare the model complexity (FLOPs and the number of parameters) with state of art methods on the X-sub benchmark in \ref {table:complexity}. Compared with most models, our model has a higher accuracy while having low computational complexity and parameter quantity. Specifically, our model's FLOPS and parameter count are only 30.02\% and 15.48\% of ST-GCN, respectively. The computational complexity and parameters of our method have increased relative to EffciientGCN-B0, mainly due to adjustments in model parameters. Compared with Sp-GCN, the FLOPs the number of parameters of SpSt-GCN have only increased by 15.3\% and 9.1\% respectively, which indicates that the structural branch is very efficient.

\begin{table}[h]
\centering
  \begin{tabular}{llll}
      \toprule
      Model & Acc &  FLOPs & Param   \\
      \midrule 
      ST-GCN*\cite{yan2018spatial} & 81.5 & 16.32* & 3.10* \\
      SR-TSL*\cite{si2018skeleton} &     84.8 &    4.20*&   19.07*\\
      AS-GCN*\cite{li2019actional} &     86.8&     26.76*&   9.50*\\
      2s-AGCN*\cite{shi2019two} & 88.5 & 37.32* & 6.94* \\
      SGN\cite{zhang2020semantics} &     89.0&     -&   0.69\\
      AGC-LSTM\cite{si2019attention} &     89.2&     -&   22.89\\
      PL-GCN\cite{huang2020part} &     89.2&     -&   20.70\\
      NAS-GCN\cite{peng2020learning} &     89.4&     -&   6.57\\
      DGNN\cite{shi2019skeleton} &     89.9&     -&   26.24\\
      EfficientGCN-B0\cite{song2022constructing} & 90.2 & 2.73 & 0.29 \\
      4s-Shift-GCN\cite{cheng2020skeleton} & 90.7 & 10.0 & 2.76* \\
      DC-GCN+ADG*\cite{cheng2020decoupling} &   90.8&     25.72*&   4.96*\\
      MS-G3D*\cite{liu2020disentangling} &   91.5&     48.88*&   6.40\\
      Dynamic-GCN\cite{ye2020dynamic} & 91.5 &  -& 14.40 \\ 
      MST-GCN\cite{chen2021multi} &   91.5&     -&   12.00\\
      CTR-GCN\cite{chen2021channel} & 92.4 & 1.97 & 1.46 \\
      \midrule
      Sp-GCN(ours) & 91.1 & 4.25& 0.44\\
      SpSt-GCN(ours) & 91.6 & 4.90& 0.48\\
      \bottomrule       
  \end{tabular}\\
  *: These results are implemented by \cite{song2022constructing}
  \caption{Comparisons with SOTA methods on NTU RGB+D X-sub benchmark on accuracy(\%), FLOPs($\times 10^9$) and parameter number ($\times 10^6$), Sp-GCN has only spatial branch in GCN layer, SpSt-GCN has both spatial and structural branch in GCN layer.}
  \label{table:complexity}
\end{table}

\section{Conclusion}
In this article, we propose a Spatial-Structural Graph Convolutional Network (SpSt-GCN) for skeleton-based action recognition to better represent symmetrical human structures and edge nodes and address the issue of over-smoothing in Graph Convolutional Networks. The spatial branch aggregates information based on the topological structure, while the data-driven structural branch performs differentiation based on the similarity of edge node sequences. Our approach yields favorable results while maintaining efficiency. In the future, we will explore how to improve the flexibility of structural connection and how to extend this method to other graph recognition tasks, e.g. In this article, we have only considered the structural connectivity of edge nodes. The representation of structural connectivity between non-edge nodes is also worth exploring.

\bibliographystyle{plain}
\bibliography{Skeleton-Based_Action_Recognition_with_Spatial-Structural_Graph_Convolution.bib}

\end{document}